# Predicting drug-drug interactions using heterogeneous graph neural networks: HGNN-DDI

**Hongbo Liu[1,4,*,†], Siyi Li[2,5,†], Zheng Yu[3,6,†]**

[1]College of Information Science and Engineering, Northeastern University, Shenyang, 110819, China

[2]Lochlann Quinn School of Business, University College Dublin, Dublin, D04 V1W8, Ireland

[3]Faculty of Science, McGill University, Montre´al, H3A 2T5, Canada

[4]loren5555@sina.com
[5]siyi.li@ucdconnect.ie
[6]zheng.yu3@mail.mcgill.ca
*corresponding author
†These authors contributed equally to this work

**Abstract.** This research centers on predicting drug-drug interactions (DDIs) using a novel approach involving graph neural networks (GNNs) with integrated attention mechanisms. In this method, drugs and proteins are depicted as nodes within a heterogeneous graph. This graph is characterized by different types of edges symbolizing not only DDIs but also drug-protein interactions (DPIs) and protein-protein interactions (PPIs). To analyze the chemical structures of drugs, we employ a pretrained model named ChemBERTa, which utilizes simplified molecular input line entry system (SMILES) strings. The similarity between drug structures based on their SMILES strings is determined using the RDkit tool. Our model is designed to establish and link heterogeneous graph neural networks, taking into account the DPIs and PPIs as key input data. For the final prediction of interaction types between various drugs, we use the Multi-Layer Perception (MLP) technique. This model's primary objective is to enhance the accuracy of DDI predictions by factoring in additional data on both drug-protein and protein-protein interactions. The forecasted DDIs are presented with associated probabilities, offering valuable insights to healthcare professionals. These insights are crucial for assessing the potential risks and advantages of combining different drugs, particularly for patients with diseases at different stages of progression.

**Keywords:** Graph Neural Network, Drug-drug Interaction, Machine Learning, Heterogeneous Graph Learning.

## 1. Introduction

Drug-drug interactions (DDIs), a prominent aspect of adverse drug reactions, occur when the combined effects of multiple medications diverge from the expected results if these drugs were used separately [1]. These interactions are broadly categorized into pharmacokinetic and pharmacodynamic types. Pharmacokinetic interactions involve changes in a drug's absorption, distribution, metabolism, and excretion (ADME), thus altering its bioavailability. For instance, fluoxetine's inhibition of the enzyme CYP2D6 increases the effectiveness of metoprolol, a beta-blocker, by slowing its metabolism, while simultaneously decreasing the efficacy of codeine by preventing its conversion to morphine. On the other hand, pharmacodynamic interactions occur when drugs affect each other's actions through their mechanisms of action. An example is the way glucagon can counter the effects of beta-blockers via activation of Gs-coupled GPCR, or how ibuprofen can inhibit aspirin's effect on platelet aggregation by blocking its access to the active site of COX-1 [2-4].

The relevance of DDIs in healthcare is emphasized by their high incidence rate and the wide range of their impacts, from beneficial synergistic effects to reduced drug effectiveness, and even potentially fatal side effects [2]. This challenge is particularly acute in hospitals, where patients often receive complex medication regimes involving multiple drugs. The likelihood of encountering DDIs in such settings is significantly higher due to the variety of medications prescribed and frequent changes in dosages and drug types. Hospitalized patients, typically more critically ill than those treated as outpatients, have a diminished capacity to cope with these pharmacological challenges. A study in a major university hospital in the Netherlands reported that 27.8% of 21,277 admissions were affected by at least one DDI [5]. While hospitals have implemented electronic DDI alert systems, these can lead to 'alert fatigue', where the high volume of alerts, many of which may be of low clinical relevance, increases the risk of overlooking critical alerts. Additionally, these systems are often limited to known DDIs from existing literature, offering limited guidance on newer drugs or untested drug combinations [5]. This highlights the need for continuous research and development of more advanced and comprehensive DDI detection and management tools in healthcare settings.

Graph Neural Network (GNN) is a form of machine learning algorithm widely used in biological and chemical problems due to the ability of graphs to represent information whose structure and connections cannot be described by simple ordering, such as the connectivity and spatial relationship of components within small molecules and macromolecules. According to Gligorijevic´ et al. (2021), they used multiple machine learning techniques for protein function prediction, such as CNN-based DeepGO, BLAST baseline, LSTM language model, and GCNs (Graph Convolutional Networks) in the form of DeepFRI, which are used for annotation transfer and to extract features from proteins, taking into account their graph-based structure of interconnected residues represented by contact maps [6].

The systematic and relatively predictable nature of drug interactions has led to the use of many machine learning algorithms, including GNNs, to tackle the problem of DDI prediction [2]. As all drug interactions lie in the binding and reaction of drug molecules with enzymes and receptors as well as that between enzymes and receptors, knowledge graphs incorporating these as nodes and their interactions as edges would be extremely valuable for DDI prediction. The use of graph neural networks to learn the topological neighborhood representation of drugs in the knowledge graph for DDI prediction was first pioneered by Lin et al (2020), who have achieved the extremely high F1 of 95.7% on binary classification of whether a given DDI exists [7]. Chen et al. combined 2D drug molecular graphs and large-scale knowledge graphs in their MUF- FIN model, and was able to achieve performance surpassing all previous models on both binary- class, multi-class and multi-label DDI prediction [8]. The use of language models to extract and incorporate textual information into DDI prediction by GNN is also a fruitful approach, as shown by the 3DGT-DDI architecture by He et al [9]. Their model represents the 3D structure of each drug as a 3D graph and uses SchNet to extract their 3D characteristics and combines this information with a SCIBERT-based text feature extraction model, giving an 84.48% macro F1 score in the DDI Extraction 2013 shared task dataset.

In this study, we introduce an innovative model that integrates graph attention neural networks with advanced pre-trained language models. This model is designed to predict drug-drug interactions (DDIs) effectively, utilizing the chemical structures of drugs, protein amino acid sequences, and a molecular biology knowledge graph. Our knowledge graph is a heterogeneous graph comprising two node types - drugs and proteins - and three edge types, including drug-drug, drug-protein, and protein-protein interactions, plus an edge representing similarity between neighboring nodes. For input, our model takes the chemical structure of drugs in SMILES format. It then employs edge prediction to determine the potential DDI edges originating from a given drug node, along with the likelihood of each edge's existence. We leverage the pre-trained language model CHEMBERTa to extract features from the SMILES representations, forming the drug node embeddings. For protein node embeddings, we utilize ESM-1b, which extracts features from protein amino acid sequences [10]. The similarity measure between feature vectors of two distinct drugs serves as a basis for the edge weight, indicating chemical similarity. Edges with weights below a predetermined threshold are not included in our graph.

Our model stands out due to its lightweight and dynamic nature. It allows for continual enhancement of DDI prediction accuracy as more drug-protein interaction information is fed into the system. This adaptability enables the model to be quickly retrained with new data. Such features make our model a valuable tool for augmenting current literature-based DDI alert systems in hospitals, especially for new and less-researched drugs and combinations. Its capability to evolve with emerging data ensures that our alert predictions remain accurate and up-to-date.

## 2. Methods and Materials

### 2.1. Datasets

#### 2.1.1. DrugBank
In 2018, the DrugBank Multi-Typed DDI dataset was curated for the first time as the Gold Standard DDI dataset for training the DeepDDI model [11]. The dataset consists of 86 types associated with at least five drug pairs. It contains 192,284 DDIs from 191,878 drug pairs. It also clearly labels each drug with the DrugBank ID SMILES string expressions corresponding to it [12].

#### 2.1.2. PrimeKG
PrimeKG is a comprehensive medical heterogeneous knowledge graph that connects over 100,000 nodes across ten types, such as drugs and proteins [13]. Each node has been labeled with clinical descriptors from trustworthy medical data sources, and it is connected by 29 types of edges containing over 4 million relationships.

The data extraction process involves identifying them with DrugBank IDs based on the protein sequence. Then, the ID of each drug should be mapped to the corresponding DrugBank ID in order to identify the same drugs in both datasets. By employing this method, it is possible to improve the accuracy and applicability of drug-drug interaction forecasts.

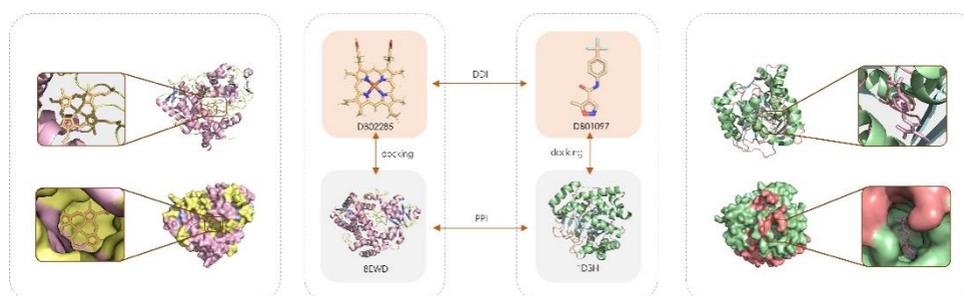

**Figure 1.** Meta path for Drug-Drug Interaction.

Drug-Protein Interaction and Protein-Protein Interactions, which illustrates that there may be docking happening between drug-protein interactions.

As illustrated in Figure 1, the utilization of drug-protein interactions and protein-protein interactions can improve the efficacy and accuracy of drug-drug interaction prediction. The establishment of a chain association has the potential to affect the molecular interactions.

*2.2. Data Analysis*

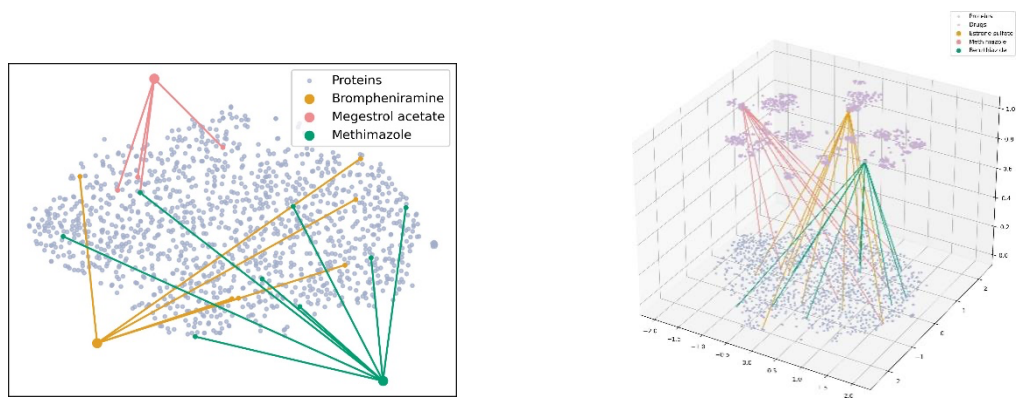

(a) 2D relations between three types of interactions.  (b) 3D relations between three types of interactions.

**Figure 2.** Meta path in dataset

We used the DrugBank Multi-Typed DDI (TDcommons) dataset to predict potential drug-drug interactions. In addition, the Drug-Protein and Protein-Protein interactions data extracted from the PrimeKG database are used as a supplement for improved prediction. As portrayed in Figure 2a and Figure 2b, a brief illustration of DDIs, DPIs, and PPIs is provided. Brompheniramine, megestrol acetate, and methimazole were chosen to demonstrate drug and protein interactions. In the meta path, both Brompheniramine and megestrol acetate can interact with a specific protein. This implies that one drug has the potential to influence the other drug by altering the protein's quantity.

The dataset contains a comprehensive list of 1,706 unique drug types and 1,544 specific protein types. Based on the calculations performed, the analysis reveals 191,808 distinct types of drug-drug interactions, 12,801 distinct types of drug-protein interactions, and 6,793 distinct types of protein-protein interactions.

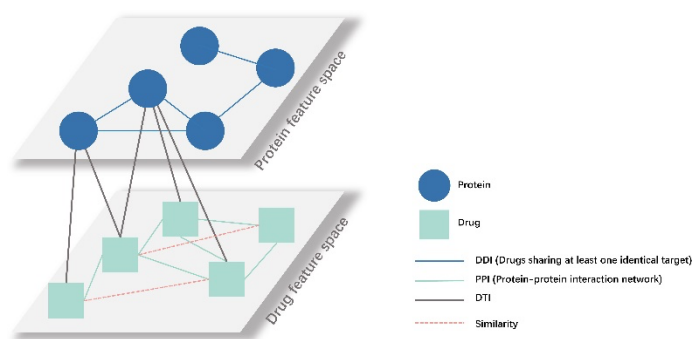

**Figure 3.** The architecture of the heterogeneous graph for two nodes and four edges with Drug-Drug interactions, Drug- Protein Interactions and Protein-Protein interactions.

The heterogeneous graph is constructed as the general structure depicted in Figure 3, which is a potent knowledge representation that integrates three distinct datasets: DDIs, DTIs, and PPIs. Traditional methods for predicting DDI types rely heavily on analyzing of individual drug pairs. However, the heterogeneous graph permits us to model complex interactions and dependencies between drugs, targets, and proteins within a unified framework. GNNs can effectively capture these interconnected relationships, which leads to more accurate data representations.

Table 1. Six types of interaction types which are grouped based on their features

| Numbers | Type |
|---|---|
| 1 | Absorption |
| 2 | Distribution |
| 3 | Metabolism |
| 4 | Excretion |
| 5 | Toxicity |
| 6 | Effects (not necessarily bad) |

Based on the analysis of a dataset containing 86 types of Drug-Drug Interactions, it has been determined that the interactions can be effectively categorized into six distinct groups based on their detailed information for their different impact on various human body systems. The types of interactions are shown in Table 1: absorption, distribution, metabolism, excretion, toxicity, and effects. Among the interaction types 46, 47, and 73, the type 6 interactions, which relate primarily to the effects of the drugs, occur most frequently. This emphasizes the importance of understanding the potential effects of drug combinations on patients, as these interactions can substantially impact their health outcomes. In addition, Figure 4 visually represents the distribution of Drug-Drug Interactions across these six groups. It reveals that type 6 interactions are the most prevalent among the various groups, highlighting the need to comprehend the impact of drug combinations on patients.

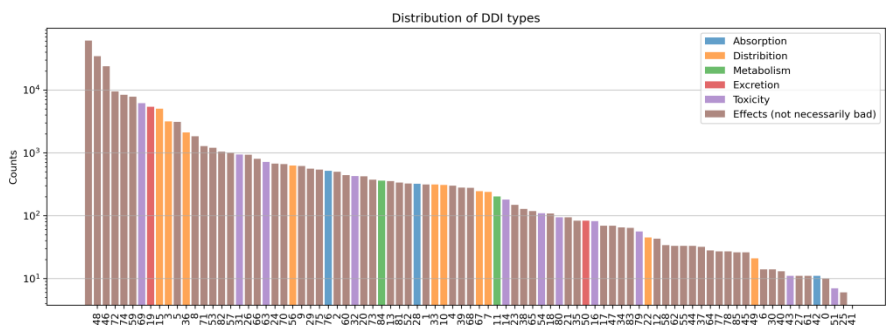

**Figure 4.** Distribution of Interaction Types of Drug-Drug Interactions.

Figure 4 shows the Type 49 interactions appear most frequently then others and the whole data distribution is certainly sparse.

*2.3. Model Architecture*

*2.3.1. ChemBERTa*
ChemBERTa, a BERT (Bidirectional Encoder Representations from Transformers) language model, has been pre-trained on 77M data to analyze and manipulate chemical and molecular data to forecast chemical properties and generate molecules [14]. It encodes SMILES strings to hold semantic information about drugs, which can provide more informative content. ChemBERTa is useful in drug discovery and other chemistry tasks because it can understand chemical language and context.

*2.3.2. ESM-1b*
ESM-1b, a general-purpose protein language model, can predict protein structure, function, and other properties from sequences [15]. It is capable of translating protein sequences into language, which provides a contextualized protein sequence representation that captures important features and patterns. Our model uses ESM-1b to create a vector that captures the protein's unique properties. The format can be used for protein-protein interaction analysis and assist in predicting drug-drug interactions.

*2.3.3. RDKits*
RDKits is free molecular modeling and cheminformatics research software, which provides many molecular structure manipulation and analysis tools [12]. It can help us understand complex molecule interactions, which is useful in predicting drug-drug interactions, where molecules with information to targets are needed for Morgan Fingerprint which will be used to calculate the similarity of molecules in our approach.

*2.3.4. Graph Neural Network (GNN)*
Graph Neural Networks (GNNs) are a category of deep learning architectures designed specifically for analysing graph-structured data, such as molecular graphs [16]. GNNs facilitate the acquisition of non-local connections and higher-level structural information within graph structures, which allows a more comprehensive understanding of medication interactions.

The Graph Convolutional Network (GCN) and Graph Attention Network (GAT) models are applied in this paper. The GCN is a graph neural network that collects and integrates local neighborhood data from a node in a graph structure, which can graph medicine relationships for DDI prediction. Drugs are nodes, and their interactions are edges in this graph. It uses a predetermined graph convolution operation to iteratively improve node representations by integrating adjacent nodes [17]. The GAT uses attention mechanisms to capture adjacent significance during information aggregation, which facilitates nodes to assign different attention weights to their neighbors, enabling them to focus on more relevant and informative neighbors. The attention mechanism in GAT makes it possible for the model to capture the diverse significance of neighboring nodes [10].

*2.3.5. Link Prediction*
Link prediction is a machine learning and graph analysis specialization that forecasts the probability of a connection between two nodes within a given graph. In the context of DDI prediction, the link prediction technique can be used to identify potential drug-drug interactions [18]. It assists in developing new medicines by analyzing the complex relationships between numerous molecular attributes.

*2.4. Overview of HGNN-DDI Model*

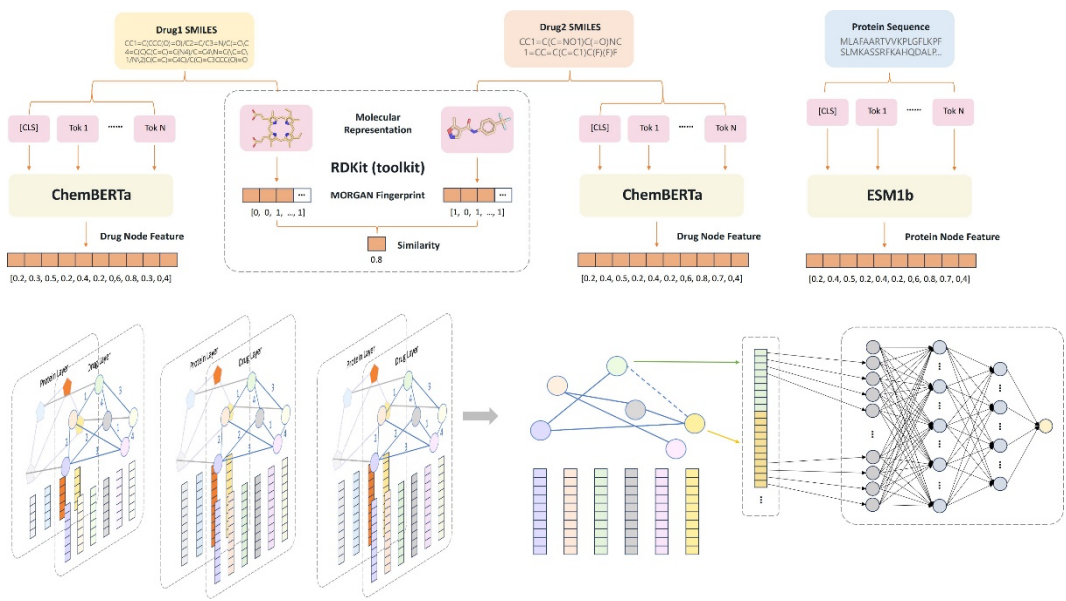

**Figure 5.** Model Architecture of the HGNN-DDI for Drug-Drug Interactions Prediction

When building our model, we relied on the Deep Graph Library (DGL), a powerful framework tailored for graph neural networks, which enabled us to effectively tackle the complexity of DDIs [19]. The architecture of our model is shown in Figure 5, and the implementation of our model proceeds as follows:
1. Input Processing: The SMILES representations of drugs are put into the ChemBERTa model for preprocessing. The chemical structure of proteins is input into the ESM-1b model for feature extraction and processing.
2. Graph Construction: The preprocessed data is represented as a graph, with drugs and proteins as nodes and their interactions and similarity as edges. This graph-based representation generates a network that elegantly summarises the interactions between different drugs and proteins.
3. Graph Convolutional Network (GCN): A three-layer GCN is applied to incorporate this knowledge effectively. As the model goes through multiple graph convolution layers, it gradually refines and consolidates its understanding of DDI interactions, resulting in representations that are abundant in features.
4. Graph Attention Network (GAT): It allows each node to assign different attention weights to each neighbor, enabling the node to focus on more significant and instructive neighbors.
5. Link Prediction and Multi-Layer Perception (MLP): The GNN-derived enriched feature vectors are seamlessly incorporated into a Multi-Layer Perception (MLP), which is well-suited for transforming the features into a classification scheme. It can capture complex nonlinear relationships, making it suitable for accurately predicting interactions between various drugs.

## 3. Experiments and Results

Through the methods mentioned above, we conducted training and testing of the model on the meticulously curated dataset with DDI, DPI, and PPI interactions. Moreover, to improve the ability of predictions for unknown drugs, we supplied the model with similarity information calculated by SMILES. We established that edges representing the two nodes are similar for molecular pairs exhibiting a similarity of over 0.7.

$$\text{accuracy}_i = \frac{TP_i + TN_i}{TP_i + TN_i + FP_i + FN_i} \quad (1)$$

$$\text{precision}_i = \frac{TP_i}{TP_i + FP_i} \quad (2)$$

$$\text{recall}_i = \frac{TP_i}{TP_i + FN_i} \quad (3)$$

$$F1_i = \frac{2 \times \text{precision}_i \times \text{recall}_i}{\text{precision}_i + \text{recall}_i} \quad (4)$$

$$W_i = \frac{N_i}{\sum N_i} \quad (5)$$

$$\text{accuracy}_{weighted} = \sum W_i \cdot \text{accuracy}_i \quad (6)$$

$$\text{precision}_{weighted} = \sum W_i \cdot \text{precision}_i \quad (7)$$

$$\text{recall}_{weighted} = \sum W_i \cdot \text{recall}_i \quad (8)$$

$$F1_{weighted} = \sum W_i \cdot F1_i \quad (9)$$

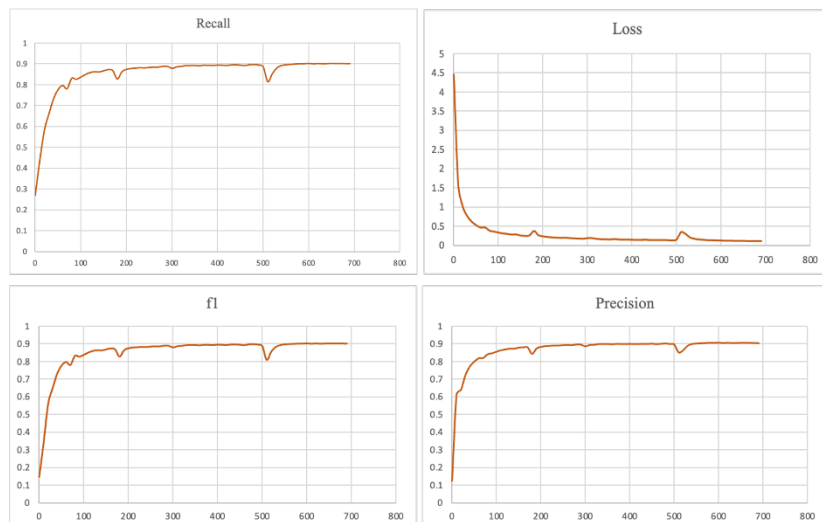

**Figure 6.** The f1 score, loss, precision and recall of HGNN-DDI model

During the testing phase, negative samples constituted 10% of the entire test set, obtained by randomly sampling negative edges from the complete test graph. Commonly used evaluation metrics such as accuracy, precision, recall, and f1 score are used to assess the performance of the model, which equations are shown above. Throughout the model training process, the parameter curves exhibited the trends shown in Figure 6.

As mentioned above, we have tried both HGAT-DDI and HGCN-DDI models, and we apply the machine learning evaluation metric of F1, Recall, Precision, and Accuracy to measure the performance of our model. By searching the relevant models that applied different models, we compare our model with them, for we are all based on the same dataset. The HGAT-DDI model is with the attention mechanisms, which means it can focus on more significant models. It is suggested to perform better that the HGCN-DDI model.

**Table 2.** The performance of different algorithms on the Drugbank dataset

| Model | Method | | F1 | Recall | Precision | Accuracy |
|---|---|---|---|---|---|---|
| GNN on DDI model | GAT | | 84.72 | 85.07 | 84.36 | 84.65 |
| | GCN | | 85.02 | 86.29 | 83.78 | 84.79 |
| | GraphSAGE | | 84.18 | 85.5 | 82.9 | 83.93 |
| GE on DDI graph | Node2Vec | | 79.63 | 78.35 | 80.95 | 88.73 |
| | DeepWalk | | 78.8 | 78.27 | 79.44 | 88.64 |
| GNN on Heterogeneous graph | GAT | | 85.68 | 86.28 | 85.99 | 86.06 |
| | GCN | | 88.28 | 89.01 | 85.49 | 89.68 |
| | GraphSAGE | | 87.79 | 88.31 | 86.94 | 85.59 |
| Our model based on Heterogeneous graph | HGCN | 6 classes | 96.91 | 94.23 | 97.01 | 96.86 |
| | | 86 classes | 90.16 | 88.45 | 90.88 | 90.47 |
| | HGAT | 86 classes | 90.35 | 90.12 | 89.68 | 90.21 |

We present two matrices: one displays the outcomes for a dataset categorized into eighty-six distinct groups, and the other shows the results for the dataset segmented into six different categories. The six categories are defined as follows: Class1 pertains to absorption processes, Class2 to distribution, Class3 to metabolism, Class4 to excretion, Class5 to toxicity, and Class6 to neutral effects, which don't necessarily indicate adverse outcomes. On the other hand, the eighty-six categories are based on varying degrees of severity in drug-drug interactions.

Our experimental findings indicate that the classification accuracy was markedly higher for the dataset when it was divided into the broader six classes, as opposed to the more detailed eighty-six class structure. In the context of the six-class system, the distinctions between each group are more pronounced, thus facilitating the model's ability to differentiate between the classes. However, in the 86-class scenario, the nuances between some of the categories are less distinct, posing a challenge for the model in achieving precise classification.

Contrary to our expectations, the HGAT model showed minimal improvement compared to the HGCN model. In the experiment, the Heterogeneous Graph Convolutional Neural Network model achieved a maximum F1 score of 0.969, with precision, recall, and accuracy all exceeding 90%. In the dataset with 10% negative samples, the HGAT model's performance was nearly identical to HGCN, with all metrics hovering around 90%. As shown in Table 2 above, it compares performance metrics between our algorithm and other algorithms on the same dataset. It is obvious that our model is performing better than our models with the combination of Protein-Protein interactions and Drug-Protein Interactions as well as the Heterogeneous graph to link these complex relationships together and demonstrate it clearly.

Based on this result, the current bottleneck affecting the algorithm's performance is believed to lie within the dataset. Additionally, using the simple MLP classifier without any tricks for classification using embedding vectors might also influence the algorithm's performance, potentially masking the distinctive characteristics of the GNN (Graph Neural Network).

## 4. Conclusion and Discussion

In conclusion, we proposed a novel deep learning framework that combines large-scale pre-trained language models with graph neural networks to integrate information inherent to the structure of drugs and proteins and the topological information of their position and interaction in the knowledge graph for multi-class DDI predictions. We have demonstrated that the efficacy of our model is comparable to that of state-of-the-art models that combine both knowledge graphs and drug molecular graphs despite being somewhat simpler in architecture and requiring less training time. Moreover, the high performance of our simple model is evident from the negligible F1 improvement of the HGAT model over it. Our confusion matrix also shows that we have successfully overcome the problem of promiscuously interacting drugs, reducing model accuracy that plagues relatively well-performing traditional machine learning algorithms such as Decision Trees.

## 5. Future Goals

Currently, in order to improve the accuracy of drug-drug interaction prediction, heterogeneous graphs are being exploited by GNN to improve drug-drug interaction prediction. In addition, we have incorporated the examination of Protein-Protein and Protein-Drug interactions into our methodology to help the prediction of drug-drug interactions. After performing a preliminary analysis utilizing SMILES representations for drug-drug interactions (DDI) and sequence for protein-protein interactions (PPI), we acquire the characteristics of drugs and proteins. By including DPIs and PPIs, we achieve our goal of overcoming the limitations of the original dataset. Future research objectives include integrating data containing molecular 3D conformations. The addition of additional information has the potential to increase the comprehensiveness of molecular structure and interaction types, thereby enhancing the accuracy and interpretability of our models. The use of SMILES strings data only may contain limited information, which should be improved in the future, and additional research should be conducted. To increase the exhaustiveness of the experiment, future steps, including validation, hyperparameter tuning, and further evaluation, will be taken.

## Author Contribution

Hongbo Liu: Set up training server, Implement training and network structure, Feature Extraction, Problem solving in coding and working environment.

Siyi Li: New model ideas, Literature Review, Visualize the dataset and model architecture, Model Construction of GAT, RDKit similarity Calculation.

Zheng Yu: Dataset Organization and collection, Developing toolkits of operating datasets, Molecular similarity Calculation, SNAP dataset Investigation.

**Acknowledge**

**Appendix A. Explanation of Measurements of Results**

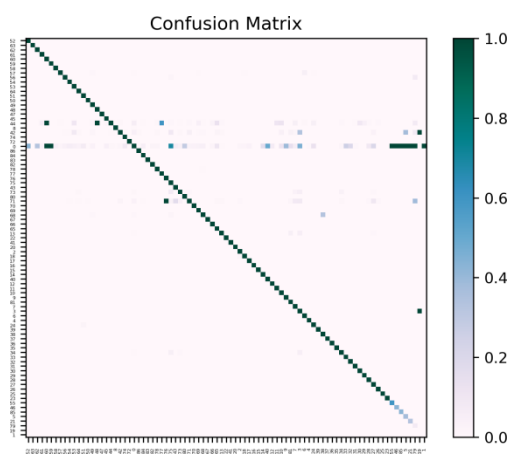

**Figure A.1.** The confusion matrix of HGCN model

For the prediction results, we plotted the confusion matrix in Figure A.1. The vertical axis in the graph represents the true values, while the horizontal axis represents the predicted values. It can be observed that the model's predictions are not accurate enough for category 0, which corresponds to cases where there is no interaction between two drugs.

However, there has yet to be much adjustment to the number of negative samples in the training set. Only 10% of negative samples exist in the training set. These negative samples did not provide enough information to the model to predict the absence of interactions between two nodes. As a result, a large number of negative samples in the test set were predicted as other categories. Furthermore, among the 1706 drug interactions, 93% (2716922/2908730) have no relationship, which is a significant proportion.

Consequently, the test set contains a wide variety of negative samples, exacerbating this effect. This issue can be alleviated by increasing negative sampling in the training set. The model can learn more about the absence of relationships by augmenting the number of negative samples in the training set, thereby improving prediction performance.

Additionally, the classification accuracy for the categories at the end of the axes is noticeably lower than the average. This is due to the imbalance of samples in the dataset, where the vast majority of the later categories have very few instances, sometimes only a few dozen or even less. Such categories make it challenging to learn the interaction features, resulting in poor prediction performance. This phenomenon can be alleviated through data set balancing methods. However, if substantial improvement in predicting these categories is desired, seeking new datasets to complement the existing data is necessary.

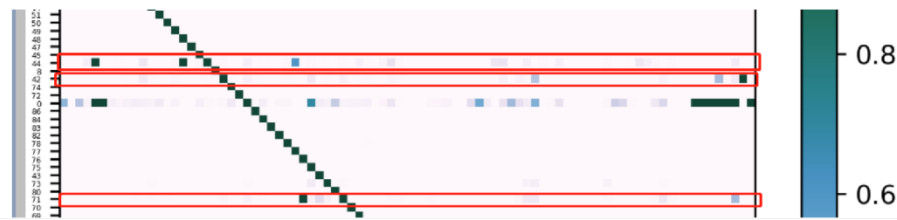

**Figure A.2.** Results of type 44, 42 and 71

As shown in Figure A.2, Drugs in Class 44 (Drug A may increase analgesic activities of Drug B), 42(Drug A may decrease vasoconstricting activities of Drug B), and 71(Drug a may increase myopathy rhabdomyolysis activities of Drug B) was seen to have a relatively higher probability of being mispredicted into other classes. This may be because of the high amount of DDIs being classified into these types in the dataset, with these three classes being the most populated classes. Therefore, these interactions may be caused by diverse mechanisms that are difficult to summarize into any one fixed pattern, while similar mechanisms mostly cause interactions in any other class.

To illustrate the impact of negative samples in the test set on the results, we conducted additional experiments using different negative sampling strategies. Specifically, we performed two variations: one without adding any negative samples and the other with all negative samples added. The confusion matrices for these experiments are shown in Figure A.3.

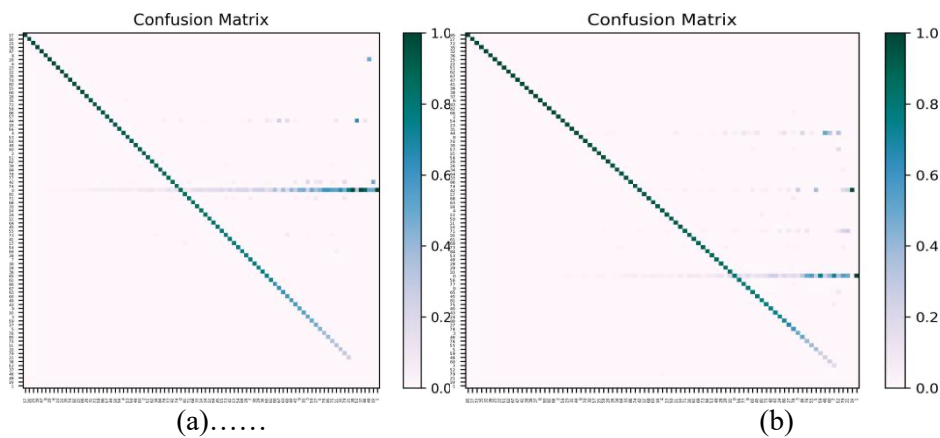

(a)……                                      (b)

**Figure A.3.** The confusion matrix of different test sets.

On the left is the confusion matrix obtained from testing on a test set that contains all 400,000 negative samples from the test graph.On the right is the confusion matrix obtained from testing on a test set that does not include any negative samples.

From the graph, it can be observed that the category "No Interaction" exhibits a lower prediction accuracy, particularly when it is confused with categories having fewer samples. When the dataset includes a significant number of negative samples, the model's F1 score is 0.871, accuracy is 0.854, precision is 0.920, and recall is 0.854. When all negative samples were removed from the dataset, the problem was transformed from link prediction to a graph edge classification problem. For this type of problem, our model achieved an F1 score of 0.950, accuracy of 0.936, precision of 0.963, and recall of 0.936. This demonstrates that graph networks can effectively learn features and connections between nodes. If further adjustments are made to the dataset, the algorithm's predictive performance is expected to improve further.

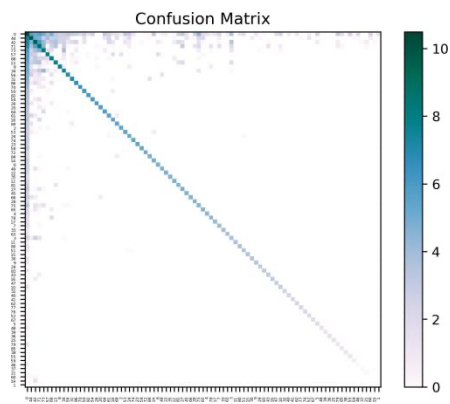

**Figure A.4**. The confusion matrix of HGAT model

Building upon the previous experiments, to demonstrate that the algorithm's predictive performance bottleneck is a result of the dataset, we conducted additional experiments by replacing the network architecture. In the earlier experiments, our graph network utilized a 3-layer GCN, a fundamental network structure. In the subsequent experiments, we employed the Graph Attention Network (GAT) layer. This network structure has been proven to possess superior feature extraction and information aggregation capabilities compared to GCN. The confusion matrix obtained using the GAT layer for testing is shown in Figure A.4.